# A Novel Evaluation Metric for Unsupervised Learning in AIS-Based Maritime Anomaly Detection: MADQI

Dr. Ismet Gocer *
Southampton Solent University
School of Technology and Maritime Industries
Southampton, United Kingdom
ismet.gocer@solent.ac.uk
* Corresponding Author

Prof. Zakirul Bhuiyan
Southampton Solent University
School of Technology and Maritime Industries
Southampton, United Kingdom
zakirul.bhuiyan@solent.ac.uk

Dr. Raza Hasan
Southampton Solent University
School of Technology and Maritime Industries
Southampton, United Kingdom
raza.hasan@solent.ac.uk

Dr. Shakeel Ahmad
Southampton Solent University
School of Technology and Maritime Industries
Southampton, United Kingdom
shakeel.ahmad@solent.ac.uk

**Funding Statement:** This paper is produced based on research conducted under the SEAGUARD project, a European Union Horizon Europe research and innovation project (Grant Agreement No. 101168489).

## Abstract

This paper introduces a new systematic framework for detecting anomalies in maritime Automatic Identification System (AIS) datasets. These anomalies include abnormal vessel behaviours related to speed, position jumps, time gaps, and turn angles. Although unsupervised learning algorithms such as Isolation Forest are widely used for detecting anomalous vessel movements, they often lack systematic and meaningful evaluation measures. To address this limitation, we propose a novel quality metric called Maritime Anomaly Detection Quality Index (MADQI). The prosed MADQI is a composite index designed to evaluate the anomaly detection performance of machine learning models without requiring labelled data. The proposed framework uses Haversine distance calculations to analyse AIS datasets and identify anomalies based on their spatial and behavioural characteristics. The proposed MADQI evaluation framework integrates four interconnected metrics: Anomaly Rate Consistency (ARC), Physical Plausibility Score (PPS), Score Distribution Separation (SDS), and Extreme Case Evidence (ECE). These metrics are combined through automatic normalisation using multi-chunk evaluation and adaptive scaling techniques. Experimental results on the AIS dataset show that the proposed framework achieved a MADQI score of 80.37%, demonstrating its effectiveness for unsupervised anomaly detection. In particular, the algorithm performed strongly in identifying abnormal vessel behaviour. Among the individual MADQI components, ECE and ARC achieved scores of 0.907 and 1.000, respectively, indicating excellent capability in detecting extreme anomalies and maintaining anomaly rate consistency. Meanwhile, PPS and SDS reached 0.672 and 0.636, respectively, reflecting reasonable physical validity and statistical separation between anomalous and normal cases. Overall, these results are encouraging and demonstrate that the proposed framework provides a reliable and meaningful approach for evaluating unsupervised anomaly detection in maritime AIS data.

**Graphical Abstract**

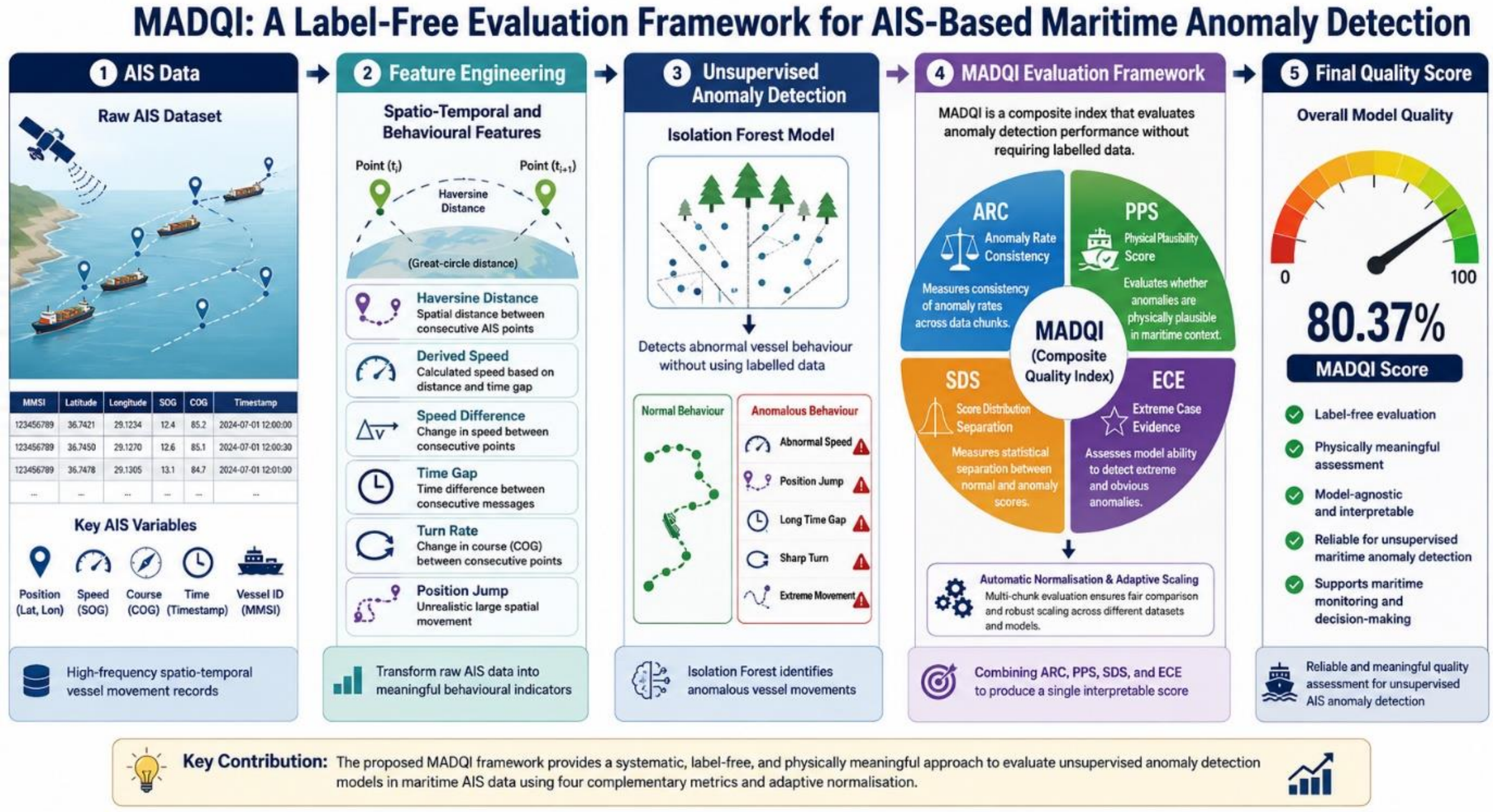

## 1. INTRODUCTION

Maritime transportation underpins more than 80% of global trade, making the monitoring of vessel movements essential for safety, security, and environmental protection. The widespread adoption of the Automatic Identification System (AIS) has enabled the continuous collection of high-frequency spatio-temporal data, providing rich information on vessel trajectories and behaviour. This has led to increasing interest in AIS-based maritime anomaly detection, particularly for applications such as illegal activity detection, collision avoidance, and maritime surveillance (Pallotta et al., 2013; Tu et al., 2018). Despite these advances, identifying abnormal vessel behaviour remains a challenging task due to several inherent characteristics of maritime data, including the lack of labelled anomaly data, high variability in maritime environments, and complex and heterogeneous vessel dynamics. These challenges necessitate the development of Artificial Intelligence (AI) models, particularly unsupervised learning approaches, to effectively detect anomalies in unlabelled data.

Isolation Forest is one of the commonly used unsupervised learning algorithms for anomaly detection in maritime security. It is an efficient method that can detect anomalies without labelled data (Liu et al., 2008). One major problem associated with this method is the absence of any benchmark evaluation metric to determine its effectiveness. The current state-of-the-art methodologies for anomaly detection in the maritime domain lack an objective and quantitative means of evaluating the accuracy of the model and the detected results, resulting in subjective analysis (Riveiro et al., 2018).

To overcome these limitations, this study introduces an innovative evaluation metric production approach, including data analytics based on AIS, feature engineering from a geospatial perspective, as well as unsupervised machine learning techniques, for assessing the performance of these AI tools. The composite Maritime Anomaly Detection Quality Index (MADQI) is a domain-specific metric that will be able to facilitate numeric, scalable, and interpretable evaluation of any anomaly detection algorithm. The suggested approach takes into consideration geospatial analytics based on Haversine (Sinnott, 1984) to detect anomalies according to vessel dynamics and categorises anomalies into five groups to make the process more structured and understandable. The main contributions are summarised as follows:

1. **A Unified Unsupervised Approach for Multi-Feature Maritime Anomaly Detection:** We propose a robust maritime anomaly detection framework which uses Isolation Forest to work successfully without any labels. This technique involves both spatial (using Haversine distance) and behavioural characteristics to be highly scalable on large amounts of data provided by AIS devices.

2. **Structured Maritime Anomaly Taxonomy:** The proposed taxonomy involves five types of maritime anomalies, considering both spatial and behavioural aspects. The structured approach allows for easy detection of vessel anomalies based on our proposed taxonomy.

3. **The Maritime Anomaly Detection Quality Index (MADQI):** A completely new quality measure for evaluating unsupervised maritime anomaly detection algorithms. MADQI incorporates three main components: spatial, temporal and behavioural. Thus, this is a novel index aimed at solving the problem of lack of common evaluation metrics for unsupervised techniques.

4. **Practical Deployment Ready Anomaly Detection and Geospatial Visualisation Tool:** The proposed methodology is engineered for seamless integration into real-time maritime surveillance and security frameworks. To enhance operational interpretability, detected anomalies are autonomously projected onto interactive geospatial maps using the Folium library. This automated visualisation layer serves as a robust decision-support platform, enabling maritime authorities to intuitively monitor vessel trajectories and identify potential threats. By bridging the gap between unsupervised detection and actionable intelligence, the framework offers a directly integrable solution for enhancing maritime situational awareness and proactive safety monitoring.

The key contributions outlined above are summarised visually in Diagram 1, which provides an integrated overview of the proposed framework and its core components.

**Diagram 1: Main Contributions of This Research**

Main Contributions and Novelty
Multi-Feature AIS-Based Anomaly Detection Framework
Robust framework using AIS data and Haversine-based spatial calculations
Five-Type Maritime Anomaly Taxonomy
Defines and operational-bes five distinct anomaly types
Unsupervised Detection via Isolation Forest
Applies Isolation Forest for anomaly detection without labels
First Composite Evaluation Metric for Maritime Anomaly
Introduces MADQI, a model-agnostic numeric evaluation index
Bridging the Evaluation Gap in Unsupervised Learning
Provides a numeric, scalable, and reproducible evaluation framework
Domain-Aware Multi-Dimensional Evaluation
Integrates spatial, temporal, and behavioural evaluation
First Composite Evaluation Metric for Maritime Anomaly Detection (MADQI)
Introduces MADQI, a model-agnostic numeric evaluation index
Integration with Real AIS Data and Geospatial Features
Leverages real-world trajectories and geospatial distance metrics
Scalable and Interpretable Evaluation Framework
MADQI is scalable and interpretable across large AIS datasets
Practical Applicability for Maritime Monitoring Systems
Enables real-time maritime surveillance, safety, and monitoring

The subsequent sections of the study are structured as follows: Section 2 presents the Related Work; Section 3 outlines the Methodology; and Section 4 reports the Experimental Results. Section 5 provides the Discussion, while Section 6 presents the Conclusion and Contributions. Finally, the study is concluded with Section 7, which discusses Future Work.

## 2. RELATED WORK

Data from AIS has been extensively utilised for analysing trajectories and detecting anomalies related to ship behaviour, thus providing a solid basis for investigating various aspects of vessel movements. Early research involved rule-based and statistical methods that made use of thresholds or probabilistic models for identifying abnormalities in vessel movement (Pallotta et al., 2013). Advanced studies moved on to the development of data-based methodologies with the use of machine learning methods capable of capturing more complicated relations within AIS trajectories. Clustering algorithms, along with trajectory mining techniques, are commonly used for detecting anomalies in ships' movements. Moreover, deep learning algorithms are now widely used for enhancing the ability to model nonlinear relationships within space-time processes (Tu et al., 2018; Zhao & Shi, 2019). Nonetheless, due to noise and irregularity within AIS data, anomaly detection becomes a challenging process.

Unsupervised anomaly detection techniques have received considerable interest owing to the lack of labelled data for anomalies. Among these, Isolation Forest is one of the most widely used techniques because of its computational efficiency and ability to isolate anomalies without prior assumptions (Liu et al., 2008). Other popular techniques include One-Class SVM (Schölkopf et al., 2001) and autoencoder-based deep learning models that can model the compact representation of normal data and detect anomalies (Sakurada & Yairi, 2014). Although these techniques are able to detect anomalies effectively, the main advantage of not requiring labelled data also poses a critical drawback. In particular, there is no benchmark or interpretable measure to assess the performance of these techniques.

One of the main challenges in maritime anomaly detection is the lack of ground truth labels, which makes objective evaluation hard. Some of the updated studies depend on qualitative approaches such as visual inspection of vessel trajectories or expert judgement to validate detected anomalies (Riveiro et al., 2018). Although these techniques provide domain insight, they are inherently subjective and difficult to reproduce. Furthermore, the lack of standardised evaluation metrics in unsupervised learning approaches limits the ability to compare models' performances. Academic research emphasises that the development of robust and scalable evaluation frameworks remains an open problem in the field (Lane et al. 2010).

Considering the limitations outlined above, this study seeks to fill an important research gap in this area. It presents the composite Maritime Anomaly Detection Quality Index (MADQI), a domain-specific, model-agnostic methodological approach for evaluating unsupervised anomaly detection in maritime applications. Different from previous work in this domain, MADQI offers a numeric evaluation criterion which can be easily interpreted and scaled for comparative purposes. The second contribution of this study is that the suggested framework facilitates benchmarking of different anomaly detection approaches using domain-specific information such as movement, behaviour, and dynamics of vessels.

## 3. METHODOLOGY

### 3.1 Data Description and AIS Data Representation

The US 31 March 2022 AIS data set utilised for the purpose of this research involves AIS-based real data with vessel trajectories over time. A trajectory is constituted by the movements made by vessels that have been identified by their unique MMSI.

Formally, a vessel trajectory can be defined as:

$$T^{(m)} = \{x_1, x_2, \dots, x_n\}$$

where $T^{(m)}$denotes the trajectory of vessel $m$ (identified by MMSI), and each observation $x_i$ corresponds to a timestamped AIS message. Each observation is represented in a multi-dimensional feature space:

$$x_i = \{lat_i, lon_i, sog_i, cog_i, t_i\}$$

where $lat_i$ and $lon_i$: geographical coordinates (latitude and longitude), $sog_i$: Speed Over Ground (SOG), $cog_i$: Course Over Ground (COG) and $t_i$: timestamp of the observation

This approach makes it possible to model the motion of vessels as a spatio-temporal pattern, accounting for both spatial and behavioural changes. To make sure that there is consistency in time, the AIS signals are sorted by their respective MMSIs in chronological order. This way, the dynamic characteristics, such as speed variations and direction changes, can be identified.

In addition, the feature space is structured to be capable of incorporating aspects related to space, time, and behaviour. In this regard, the combination of multiple features will allow the identification of irregular activities conducted by vessels, especially when considering that they take place in such complicated surroundings as maritime environments.

### 3.2 Feature Engineering

For accurate identification of anomalies in vessel behaviour, we extract a feature set from the raw AIS data, combining the use of space, time, and velocity. Feature extraction is an important component of anomaly detection using AIS data, because position data alone is not enough to detect anomalies in movement behaviour (Lane et al. 2010; Zhao & Shi, 2019).

AIS data inherently represent sequential vessel trajectories. Therefore, observations are first grouped by vessel identifier (MMSI) and sorted in chronological order. For each observation at time\ t, lag-based features from the previous state (t-1) are used to capture movement dynamics, including latitude, longitude, and course over ground. This sequential structure enables the derivation of displacement, temporal continuity, and directional change, which are essential for trajectory-based anomaly detection (Pallotta et al., 2013; Zhang et al., 2024).

#### 3.2.1 Spatial Feature: Haversine Distance

The spatial displacement between two successive AIS signals is determined using the Haversine (Half Versed Sine) formula, which is a classic great-circle distance formula often used to consider the Earth's curvature (Sinnott, 1984). It was first used by José de Mendoza y Ríos in 1801, while James Inman coined the name “haversine” in 1835. An illustration of the Haversine formula is presented in Appendix 1. The Haversine formula takes into consideration the curvature of the earth:

$$d_i = 2R \cdot \arcsin\left(\sqrt{\sin^2\left(\frac{\Delta lat}{2}\right) + \cos\left(lat_{i-1}\right) \cdot \cos\left(lat_i\right) \cdot \sin^2\left(\frac{\Delta lon}{2}\right)}\right)$$

where $d_i$ represents the distance (in kilometres) between two consecutive positions, and $R$ is the Earth’s radius. This feature enables the detection of unrealistic spatial jumps, which are commonly associated with AIS spoofing, transmission errors, or abnormal vessel behaviour (Lane et al. 2010).

#### 3.2.2 Temporal Feature: Time Difference

The temporal gap between consecutive AIS messages is defined as:

$$\Delta t_i = t_i - t_{i-1}$$

where $\Delta t_i$ is measured in seconds.

Large temporal gaps may indicate missing transmissions, while unusually small intervals combined with large displacements may signal anomalous behaviour (Pallotta et al., 2013).

### 3.2.3 Kinematic Feature: Implied Speed

The implied speed is derived from spatial and temporal changes: $v_i^{impl} = \frac{d_i}{\Delta t_i} \times 3600$. Here speed is expressed in km/h. To ensure unit consistency, implied speed is converted into knots: $v_i^{knots} = v_i^{impl}/1.852$. This feature enables a direct comparison between reported Speed Over Ground (SOG) and actual movement derived from positional changes, thereby facilitating the detection of inconsistencies in vessel motion, which are widely recognised as indicators of anomalous maritime behaviour (Kroodsma et al. 2018).

### 3.2.4 Behavioural Feature: Speed Difference

To capture inconsistencies in vessel dynamics, we compute the difference between the reported Speed Over Ground (SOG) and the implied speed: $\Delta v_i = | SOG_i - v_i^{knots} |$. Large deviations may indicate sensor errors, spoofing, or abnormal vessel operations (Kroodsma et al. 2018).

### 3.2.5 Behavioural Feature: Turn Rate (TR)

Directional changes are quantified using the rate of change in Course Over Ground (COG): $TR_i = \frac{|COG_i - COG_{i-1}|}{\Delta t_i}$. This feature captures abrupt manoeuvres, which may be indicative of evasive actions, illegal activities, or navigation anomalies (Pallotta et al. 2013).

### 3.2.6 Feature Integration

The final feature vector for each observation is defined as: $x_i^* = \{lat_i,\ lon_i,\ SOG_i,\ COG_i,\ d_i,\ \Delta t_i,\ v_i^{knots},\ \Delta v_i,\ TR_i\}$. This multi-dimensional representation combines spatial, temporal, and behavioural characteristics, enabling a more robust identification of anomalies compared to using raw AIS features alone.

### 3.2.7 Data Preprocessing and Feature Consistency

To ensure data quality, consistency, and robustness, a comprehensive preprocessing and feature standardisation pipeline is applied. AIS data represent temporally ordered vessel trajectories. Therefore, observations are grouped by vessel identifier (MMSI) and sorted in chronological order. For each observation at time $t$, lag-based features are constructed from the previous state $(t-1)$, denoted as $(Lat_{t-1}, Lon_{t-1}, COG_{t-1})$, enabling the modelling of temporal dependencies in vessel trajectories. These features enable the derivation of displacement, directional change, and temporal continuity, which are essential for trajectory-based anomaly detection (Pallotta et al., 2013; Zhang et al., 2024).

The missing values generated through lag are filtered out to preserve uniformity. Also, care is taken about timestamp accuracy since timestamp errors are common in AIS data. Coercion techniques will be applied in addressing erroneous timestamps; missing and infinite values will also be managed. Finally, duplicate entries of the same AIS message will be eliminated.

Preprocessing is key to reducing noise and increasing the effectiveness of feature extraction in real AIS data (Kroodsma et al. 2018).

### 3.2.9 Extreme Anomaly Filtering

To operationalise the anomaly taxonomy defined in Section 3.3, a rule-based filtering mechanism is applied. An observation is classified as anomalous if the corresponding feature values exceed predefined thresholds: $\Delta v_i > \tau_v,\ d_i > \tau_d,\ \Delta t_i > \tau_t,\ TR_i > \tau_r$. In addition, composite anomalies are identified when multiple anomaly indicators are simultaneously active: $C_i = 1 \quad if \quad A_{speed} + A_{jump} + A_{time} + A_{turn} \geq 2$. Here $I(.)$ is an indicator function. The resulting extreme anomaly set is defined as: $A_{extreme} = \{\, x_i \mid any\ anomaly\ condition\ holds\ \}$.

## 3.3 Anomaly Types

Based on the derived features, five types of anomalies are defined (Pallotta e al. 2013; Tu et al. 2018; Liang et al. 2024):

**3.3.1 Speed Anomaly:** A speed anomaly ($A_{speed}$) occurs when the discrepancy between the reported Speed Over Ground (SOG) and the calculated (implied) speed exceeds the threshold ($\tau_v$). This is formally defined as: $A_{speed} = I(\Delta v > \tau_v)$. Here $\Delta v = \left|SOG - V_{implied}\right|$. Significant deviations often suggest intentional data manipulation to mask unauthorized speeds or sensor inaccuracies during high-dynamic maneuvers.

**3.3.2 Position Jump Anomaly:** A position jump anomaly ($A_{jump}$) captures unrealistic spatial displacements between consecutive AIS messages that violate vessel kinematic limits. It is defined as: $A_{jump} = I(d > \tau_d)$. Here $d$ represents the distance between $P_i$ and $P_{i-1}$. To account for the temporal context, the threshold $\tau_d$ is dynamically determined as $\tau_d = V_{max} \cdot \Delta t$, here $\Delta t$ is the time interval between consecutive reports and $V_{max}$ denotes the vessel's maximum operational speed. In practice, if $d$ exceeds this physically attainable limit, the displacement is flagged as improbable, representing a clear violation of the physical constraints inherent in maritime navigation.

**3.3.3 Temporal Gap Anomaly:** A temporal anomaly ($A_{time}$) occurs when the transmission interval $\Delta t$ between successive messages surpasses the expected reporting frequency $\tau_t$: $A_{time} = I(\Delta t > \tau_t)$. Abnormally large gaps are critical indicators of "*Dark Ship*" behaviour, where transponders may be deactivated to conceal illicit activities or entry into sensitive maritime zones.

**3.3.4 Turn Rate Anomaly:** A turn rate anomaly ($A_{turn}$) captures abrupt and improbable changes in the vessel's Course Over Ground (COG) or heading: $A_{turn} = I(TR > \tau_r)$. Here $TR$ is the calculated rate of turn. High values of $TR$ that exceed the structural and hydrodynamic limits of the vessel type are flagged as potential data errors or emergency evasive maneuvers.

**3.3.5 Composite Behavioural Anomaly:** A composite anomaly is defined when multiple anomaly conditions are simultaneously satisfied: $\boldsymbol{A_{comp} = I(A_{speed} + A_{jump} + A_{time} + A_{turn} \geq 2)}$. Here $\boldsymbol{I(.)}$ denotes the indicator function and $\boldsymbol{\tau_v, \tau_d, \tau_t, \tau_r}$ are predefined thresholds. Each anomaly indicator takes binary values in $\{\mathbf{0}, \mathbf{1}\}$, enabling direct aggregation and integration into the proposed evaluation framework.

### 3.4 Unsupervised Machine Learning Model: Isolation Forest

Isolation Forest is employed as the core unsupervised learning model for anomaly detection in AIS data. Unlike supervised approaches, Isolation Forest does not require labelled data, making it particularly suitable for maritime anomaly detection where ground truth labels are typically unavailable (Liu et al., 2008).

The method is based on the principle that anomalies are more susceptible to isolation than normal observations. It constructs an ensemble of binary trees by recursively partitioning the feature space using randomly selected features and split values. Since anomalous points are rare and differ significantly from many observations, they tend to be isolated with fewer splits, resulting in shorter path lengths within the trees.

Formally, the anomaly score of an observation $x_i$ is defined based on the average path length across all trees in the ensemble:

$$s(x_i) = 2^{-\frac{E(h(x_i))}{c(n)}}$$

where $E(h(x_i))$ denotes the expected path length required to isolate observation $x_i$, and $c(n)$ is a normalisation factor dependent on the sample size $n$. Observations with shorter average path lengths yield higher anomaly scores, indicating a higher likelihood of being anomalous.

Isolation Forest is especially efficient because of its computation effectiveness, scalability for dealing with large databases and its capability of modelling complicated non-linear patterns without assuming any probability distributions. These traits make Isolation Forest especially useful in handling big AIS data, where vessel behaviour varies considerably (Manna and Bharath,2025).

The use of the Isolation Forest algorithm will be done on the selected set of features, while the obtained anomaly scores will be utilized for anomaly detection in the following MADQI procedure.

### 3.5 Proposed Evaluation Metric: MADQI

Unsupervised anomaly detection models lack well-defined evaluation metrics such as accuracy, precision, recall, and F1-score, which are commonly used in supervised learning. As a result, it becomes difficult to objectively assess the correctness, reliability, and comparability of detected anomalies.

Isolation Forest produces anomaly scores that reflect the degree of abnormality of each observation; however, these scores lack interpretability, a standardised evaluation scale, and a domain-aware assessment framework. This limitation makes it challenging to validate results and consistently evaluate detection performance.

To address this challenge, this study defines the composite *Maritime Anomaly Detection Quality Index* (MADQI) as a weighted multi-component function designed for unsupervised anomaly evaluation. MADQI integrates multiple evaluation dimensions, capturing not only the statistical properties of anomaly scores but also the temporal, spatial, and behavioural characteristics of detected anomalies, thereby enabling a more comprehensive and interpretable assessment of anomaly detection performance.

### 3.5.1 Notation

Each AIS observation is denoted by $x_i$, and the full dataset is represented as $X = \{x_i\}_{i=1}^{N}$, where $N$ is the total number of observations. For each observation, an anomaly score $s_i$ is obtained from the Isolation Forest model, reflecting the degree of abnormality. An anomaly indicator $A_i \in \{0,1\}$ is defined such that $A_i = 1$ if observation $x_i$ is classified as anomalous, and $A_i = 0$ otherwise. Based on the anomaly taxonomy introduced in Section 3.3, each observation is further associated with five anomaly indicators: $A_{speed}, A_{jump}, A_{time}, A_{turn}, A_{comp}$.

### 3.5.2 Component-Based Evaluation

MADQI is constructed as a weighted combination of four e complementary components:

**i. Anomaly Rate Consistency (ARC):** It measures the consistency between the detected anomaly ratio and the expected contamination level specified in the model.

$$ARC = 1 - \frac{| r_{obs} - r_{exp} |}{r_{exp} + r_{obs} + \epsilon}$$

where $r_{obs} = \frac{1}{N}\sum_{i=1}^{N} A_i$ is the observed contamination level, $r_{exp}$ is the expected contamination level and $\epsilon$ is a small constant for numerical stability.

**ii. Physical Plausibility Score (PPS):** It evaluates whether detected anomalies exhibit meaningful physical deviations from normal vessel behaviour, based on derived speed, travelled distance, and temporal gap.

$$PPS = \frac{1}{3}\left(\frac{Q_{0.95}^{anom}(v)}{Q_{0.95}^{norm}(v)} + \frac{Q_{0.95}^{anom}(d)}{Q_{0.95}^{norm}(d)} + \frac{Q_{0.95}^{anom}(\Delta t)}{Q_{0.95}^{norm}(\Delta t)}\right)$$

where $v$ denotes speed, $d$ travelled distance, $\Delta t$ time gap and $Q_{0.95}$ the 95$^{th}$ percentile.

**iii. Score Distribution Separation (SDS):** It captures the statistical separation between anomaly scores assigned to normal and anomalous observations.

$$SDS = \frac{| \mu_{norm} - \mu_{anom} |}{\sigma_{all} + \epsilon}$$

where $\mu_{norm}$ is mean anomaly score of normal observations, $\mu_{anom}$ is mean anomaly score of anomalies and $\sigma_{all}$ is the standard deviation of anomaly scores over all observations.

**iv. Extreme Case Evidence (ECE):** It quantifies the extent to which the detected anomalies contain evidence of extreme behavioural deviations, particularly in spatial movement patterns.

$$ECE = \frac{Q_{0.99}^{anom}(d)}{Q_{0.99}^{norm}(d) + \epsilon}$$

where $Q_{0.99}$ denotes 99$^{th}$ percentile and $d$ spatial displacement.

### 3.5.3 Normalisation and Multi-Chunk Adaptive Scaling in MADQI

To ensure comparability across heterogeneous evaluation metrics, each MADQI component must be transformed into a bounded and dimensionless scale. The raw metric values exhibit substantially different magnitudes and units, and direct aggregation would therefore lead to dominance by high-scale components. A commonly adopted approach in the literature is the soft rational normalisation (Bishop, 2006):

$$Q_k^{norm} = \frac{Q_k}{Q_k + \kappa_k}$$

where $Q_k$ denotes the raw value of the $k$-th evaluation component and $\kappa_k > 0$ is a scale parameter controlling the rate of saturation. However, when $\kappa_k$ is selected based on central statistics (e.g., median values), the transformation tends to concentrate values around 0.5, reducing discriminative power.

#### 3.5.3.1 Multi-Chunk Evaluation for Robust Scale Estimation

To address this limitation, the present study introduces a multi-chunk evaluation strategy for robust estimation of scaling parameters. The test dataset $\mathcal{D}_{test}$ is partitioned into $M$ non-overlapping subsets (Hastie et al. 2009):

$$\mathcal{D}_{test} = \bigcup_{m=1}^{M} \mathcal{D}^{(m)}, \qquad \mathcal{D}^{(i)} \cap \mathcal{D}^{(j)} = \emptyset$$

For each chunk $\mathcal{D}^{(m)}$, the raw MADQI components are computed independently:

$$\{ARC^{(m)},\ PPS^{(m)},\ SDS^{(m)},\ ECE^{(m)}\}$$

This produces an empirical distribution of each evaluation component across different portions of the test data. This approach serves as a robustness-oriented evaluation mechanism, enabling stable and data-driven estimation of normalisation parameters from repeated subtest measurements.

#### 3.5.3.2 Adaptive Selection of Scaling Parameters

Rather than specifying scale parameters manually, the present study adopts a data-driven approach based on the empirical distributions obtained from the multi-chunk evaluation. For each component $Q_k$, a scaling parameter $\tau_k$is defined as the lower quartile (25th percentile) of its chunk-level raw values:

$$\tau_k = \text{Quantile}_{0.25}\left(Q_k^{(1)}, Q_k^{(2)}, \dots, Q_k^{(M)}\right)$$

This choice reflects a lower-bound reference level rather than a central tendency. As a result, it produces:

- moderate values exceed the reference scale more easily,
- strong-performing components are not artificially constrained,
- robustness to outliers and heavy-tailed behaviour is improved.

**3.5.3.3 Exponential Normalisation**

To overcome the centralisation effect of rational scaling, this study employs an exponential saturation function (Hastie et al. 2009; Goodfellow et al. 2016):

$$Q_k^{norm} = 1 - exp\left(-\frac{Q_k}{\tau_k}\right)$$

This transformation has the following desirable properties:

- $Q_k^{norm} \in [0,1)$,
- monotonic increasing with respect to $Q_k$,
- smooth and continuous saturation behaviour,
- no forced concentration of values around intermediate levels.

In contrast to the $\frac{Q}{Q+\kappa}$ formulation, the exponential mapping allows high-performing components to approach unity more naturally while preserving relative ordering across models.

**3.5.3.4 Application to MADQI Components**

The exponential normalisation is applied uniformly to all evaluation components:

$$\begin{aligned} ARC^{norm} &= 1 - \exp\left(-\frac{ARC}{\tau_{ARC}}\right) \\ PPS^{norm} &= 1 - \exp\left(-\frac{PPS}{\tau_{PPS}}\right) \\ SDS^{norm} &= 1 - \exp\left(-\frac{SDS}{\tau_{SDS}}\right) \\ ECE^{norm} &= 1 - \exp\left(-\frac{ECE}{\tau_{ECE}}\right) \end{aligned}$$

This unified treatment ensures consistency across components while maintaining interpretability and scale invariance.

**3.5.3.5 Final MADQI Formulation**

Following the normalisation of individual components, the overall MADQI score is computed as an aggregated measure of the transformed metrics. By mapping all components onto a common bounded scale, the aggregation becomes both meaningful and balanced, preventing any single metric from disproportionately influencing the final score due to differences in magnitude and units. Formally, the MADQI score is defined as:

$$MADQI = \sum_{k=1}^{K} w_k \cdot Q_k^{norm}$$

where $Q_k^{norm}$ represents the normalised value of the $k$-th evaluation component and $w_k \in [0,1]$ denotes its corresponding weight, subject to the constraint: $\sum_{k=1}^{K} w_k = 1$.

In this study, equal weights are assigned to all components to ensure a balanced evaluation across different dimensions of anomaly detection:

$$MADQI = \frac{1}{4}(ARC^{norm} + PPS^{norm} + SDS^{norm} + ECE^{norm})$$

The resulting MADQI score lies within the interval [0, 1], where higher values indicate stronger anomaly detection performance in terms of:

- consistency with expected anomaly rates (ARC),
- physical plausibility of detected behaviour (PPS),
- statistical separability between normal and anomalous observations (SDS),
- and evidence of extreme behavioural deviations (ECE).

For interpretability, the score is additionally expressed on a percentage scale: $MADQI_{100} = 100 \times MADQI$. Higher values of $MADQI_{100}$, particularly those approaching 100, indicate stronger anomaly detection performance and a higher explanatory power of the model in capturing meaningful anomalous behaviour.

**3.5.3.6 Integrated Evaluation Framework**

The proposed MADQI formulation is embedded within a unified evaluation framework that combines:

- multi-chunk metric estimation
- data-driven parameter selection
- exponential normalisation

to provide a stable, interpretable, and domain-consistent evaluation index.

The use of multi-chunk evaluation enables the estimation of empirical distributions of raw component scores across different portions of the test dataset, ensuring robustness against data-specific fluctuations. Based on these distributions, normalisation parameters are selected automatically using lower-quantile statistics, thereby avoiding arbitrary parameter tuning.

Furthermore, the adoption of exponential normalisation:

$$Q_k^{norm} = 1 - \exp\left(1 - \frac{Q_k}{\tau_k}\right)$$

addresses a key limitation of conventional soft normalisation approaches. In particular, it prevents the artificial centralisation of scores around intermediate values (e.g., 0.5), allowing high-performing components to approach unity more naturally while preserving relative ordering between models.

#### 3.5.3.7 Advantages of the Proposed MADQI Framework

The proposed MADQI metric addresses a critical gap in unsupervised anomaly detection by providing a **model-agnostic and label-free evaluation framework**. Its key advantages can be summarised as follows:

- **Interpretability**: The decomposition into ARC, PPS, SDS, and ECE enables not only a single quantitative score but also diagnostic insight into different aspects of model behaviour.
- **Robustness**: Multi-chunk evaluation reduces sensitivity to specific test segments and provides stable parameter estimation.
- **Scalability**: The framework is computationally efficient and suitable for large-scale AIS datasets.
- **Domain-awareness**: The components explicitly capture maritime-specific dynamics, including vessel movement, spatial displacement, and behavioural anomalies.
- **Independence from labelled data**: MADQI enables consistent and practical evaluation in fully unsupervised settings, where ground truth labels are unavailable or unreliable.

This integration of adaptive normalisation, multi-chunk evaluation, and interpretable component design ensures that MADQI provides a theoretically grounded and practically applicable for anomaly detection quality in maritime environments. To provide an overview of this evaluation framework, the overall MADQI pipeline is illustrated in Diagram 2.

**Diagram 2: MADQI Pipeline**

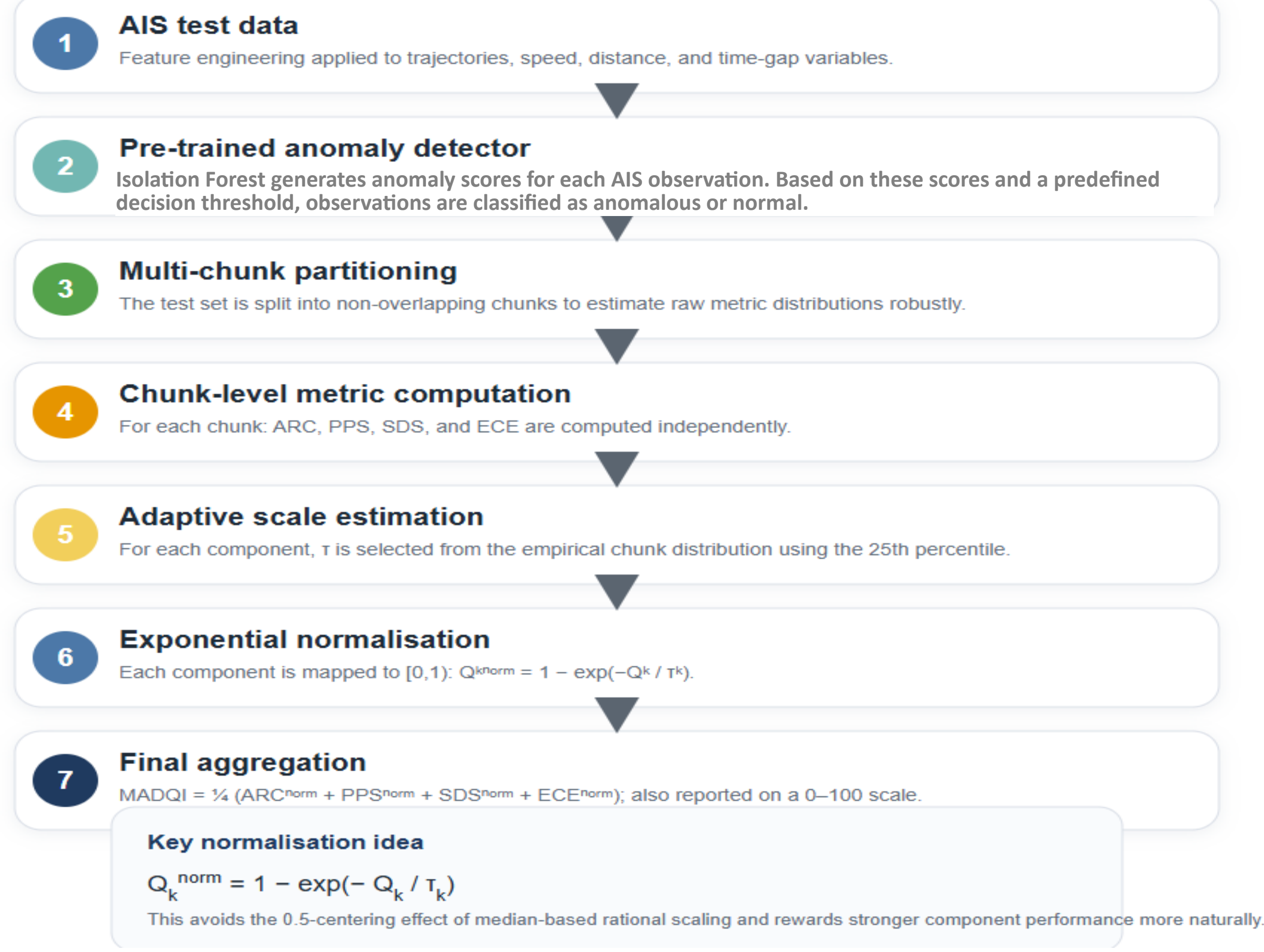

**Diagram 2** illustrates the end-to-end MADQI evaluation process, highlighting how anomaly scores are transformed into interpretable quality measures through multi-chunk analysis, adaptive scaling, and exponential normalisation, ultimately leading to a robust and comparable composite index.

## 4. EXPERIMENTAL RESULTS

This section presents the experimental evaluation of the proposed anomaly detection framework, including the dataset, model configuration, and detection settings.

### 4.1 Setup

The experimental setup is described in terms of the dataset, model parameters, and contamination level used in the analysis.

#### 4.1.1. Dataset description

In this study, we used Automatic Identification System (AIS) data of the United States on 31 March 2022 from NOAA (2022). The dataset contains a total of 1,048,576 observations, representing temporally ordered vessel movements.

Each observation includes key navigational and vessel-related attributes such as Maritime Mobile Service Identity (MMSI), timestamp, geographical coordinates (latitude and longitude), Speed Over Ground (SOG), and Course Over Ground (COG), along with additional vessel characteristics (e.g., vessel type, dimensions, and status).

Among them, the core features are positional (latitude, longitude), temporal (timestamp), and kinematic variables (SOG, COG), which form the basis for the derived spatial, temporal, and behavioural features used in anomaly detection.

#### 4.1.2 Model parameters

The Isolation Forest model was employed as the core anomaly detection method. The model was configured to ensure a balance between detection sensitivity and computational efficiency on large-scale AIS data.

First, the number of trees ($n_estimators$) was set to 100, providing stability of anomaly score computations while being relatively fast. Second, the contamination parameter, which determines the expected proportion of anomalies in the dataset, was chosen at the value of 0.001 (0.01%); reflecting the assumption that anomalous vessel behaviour is relatively rare in real-world maritime traffic. This parameter directly affects the decision threshold used to distinguish between normal and anomalous observations. A lower contamination level results in stricter anomaly detection, reducing false positives while focusing on the most extreme irregularities. Third, we decided to fix the random state (set at the value of 42). Lastly, the model was allowed to run in parallel to save time on calculations.

### 4.2 Anomaly Detection Results

This section presents the main anomaly detection outcomes obtained from the Isolation Forest model. The results are examined from three complementary perspectives: the statistical distribution of anomaly scores, maritime anomaly detection quality index (MADQI) and the spatial distribution of detected anomalies on maritime maps.

### 4.2.1 Anomaly Scores

The Isolation Forest model was applied to the first 500,000 temporally ordered AIS observations, selected from the full dataset containing 1,048,576 records. This subset was used consistently for both model training and anomaly detection to preserve methodological coherence and ensure computational efficiency.

An anomaly score was calculated for each observation, which indicated the level of deviation from normality in the new feature space created by the model. The lower the anomaly score, the more anomalous the observation, whereas a higher score implied that the vessel's behaviour was normal. In total, 498 observations were found to be anomalous, constituting about 0.1% of the analysed dataset.

In order to further differentiate the most severe cases, a secondary filtering step was adopted to detect the presence of extreme anomalies. The approach uses a combination of modelling-based ranking and domain-based rule-based ranking. Initially, the lower tail of the distribution of the anomaly scores is selected using the lower quantile threshold criterion (bottom 2% of the anomalies identified). Further constraints that include:

- **i.** Large discrepancies between reported and implied speed
- **ii.** Abnormal spatial displacements (position jumps)
- **iii.** Unusually large temporal gaps between AIS messages
- **iv.** Excessive turn rates
- **v.** Unrealistically high implied speeds

An observation was classified as an extreme anomaly if it satisfied at least one of these domain-specific conditions after passing the score-based filtering stage.

As a result, 10 observations were identified as extreme anomalies, representing the most severe and operationally relevant irregularities in vessel behaviour. This percentage is based on the contamination rate and thus does not represent any estimation based on actual occurrences. Rather, the validation process revolves around the behavioural accuracy and interpretation of these anomalies. On the whole, it can be concluded that the algorithm performs well at recognising patterns of vessel movement as opposed to anomalies, while the outlier elimination process helps in isolating critical anomalies.

### 4.2.2 Spatial Visualisation of Detected Anomalies

In order to complement the score-based analysis of anomaly detection, anomalies and extreme anomalies were visualised on an interactive Folium map. The map visualisation helps assess intuitively if the anomalies represent real maritime irregularities, such as sudden position change, abnormal movements, or geographical clustering of abnormalities. The anomaly points not only indicate a location but also include context.

More precisely, during the interaction process with the map, the user can click on each point representing an anomaly and obtain related information such as anomaly scores, computed kinematics features, and even human readable explanation for each extreme anomaly, which includes reasons for the identification of the anomaly (e.g., speed irregularity, position jumps, time gaps, and high turn rates). An example screenshot of the interactive map is shown to demonstrate the spatial distribution of identified anomalies. The visualisation part acts as a qualitative validation step before performing any quantitative validation according to the MADQI approach. A representative snapshot of the interactive map is shown in Map 1.

**Map 1: Interactive Anomaly Detection Map**

Anomalies in Map 1 are colour-coded by severity, while extreme anomalies are highlighted and annotated with explanatory behavioural indicators.

### 4.3 MADQI Evaluation

To quantitatively measure the performance of the anomaly detection task without the presence of label data, the Maritime Anomaly Detection Quality Index (MADQI) was implemented. The MADQI index enabled a comprehensive analysis of different factors through incorporating complementary aspects of evaluation into one measurement, which were obtained based on anomaly scores and behaviour features.

To testing the model, a sample of 500,000 observations is chosen, which corresponds to the second 500,000 segment of data in chronological order. This approach is necessary due to the frequent updates in AIS that create numerous signals in milliseconds. Hence, for testing purposes, it would be extremely resource-consuming to perform experiments with the rest of the data. The Isolation Forest model produces anomaly scores, where lower values indicate higher degrees of abnormality. The distribution of these scores is illustrated in Figure 1.

**Figure 1: Distribution of Anomaly Scores of Test Data**

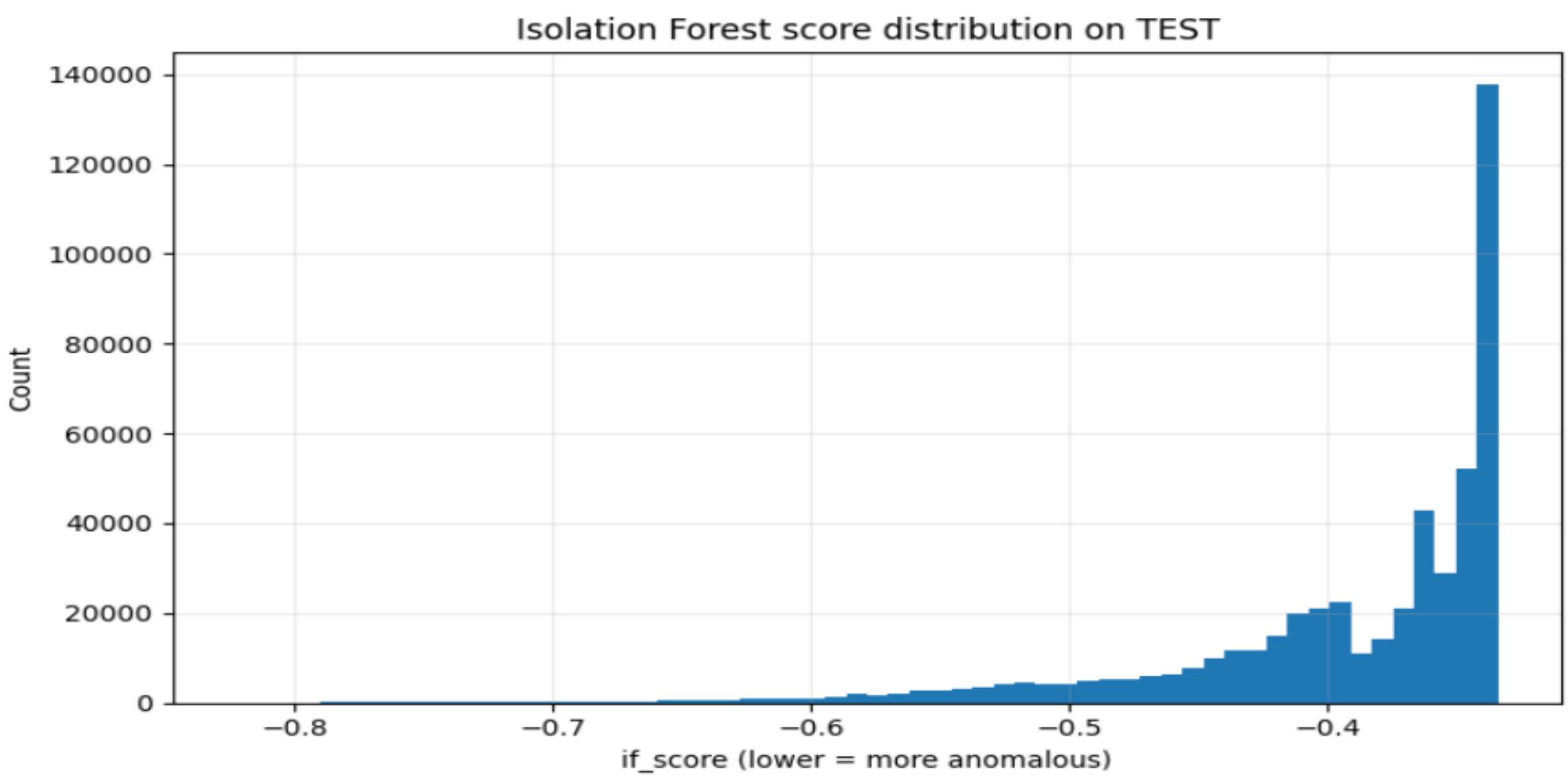

Figure 1 demonstrates the histogram for the Isolation Forest anomaly scores found on the test data set. The nearer a score is to zero, the greater the chance that it will be attributed to a normal behaviour of vessels. On the contrary, more negative numbers denote a bigger chance of the vessel exhibiting an abnormal pattern of behaviour. It is clear from Figure 2 that there is a large number of instances of normal behaviour, while there is a handful of instances in the anomalous region.

The results obtained using the metrics based on multi-chunks (5 chunks) are provided in Table 1. Every chunk is an independent segment of the test dataset that can be used to evaluate the variation in the performance of anomaly detection.

**Table 1: Chunk-level Metrics**

| n_rows | n_anom | r_obs | r_exp | ARC_raw | PPS_raw | SDS_raw | ECE_raw | chunk_id | start_idx | end_idx |
|---|---|---|---|---|---|---|---|---|---|---|
| 100000 | 39 | 0.0004 | 0.001 | 0.561 | 141.035 | 5.686 | 1529.635 | 1 | 0 | 100000 |
| 100000 | 18 | 0.0002 | 0.001 | 0.305 | 97.813 | 5.771 | 55.695 | 2 | 100000 | 200000 |
| 100000 | 2 | 0.00002 | 0.001 | 0.040 | 79.197 | 5.999 | 30.729 | 3 | 200000 | 300000 |
| 100000 | 10 | 0.0001 | 0.001 | 0.182 | 484.054 | 5.939 | 447.576 | 4 | 300000 | 400000 |
| 100000 | 69 | 0.0007 | 0.001 | 0.816 | 67.129 | 5.320 | 32.414 | 5 | 400000 | 500000 |

Evaluation results of each chunk are given in Table 1 below. These chunks are defined as a non-overlapping set within the test dataset. These results confirm that using a single test segment may lead to biased evaluation. For this reason, the next researchers should also use a multi-chunk framework to obtain a more robust and representative assessment.

Building upon the chunk-level distributions, adaptive scaling parameters ($\tau_k$) are estimated using lower-quantile statistics. The automatically selected values, including the 25th percentile, median, and 75th percentile for each component, are reported in Table 2.

**Table 2: Automatically Selected Taus**

| Component | Tau_q25 | Tau_median | Tau_p75 |
|---|---|---|---|
| ARC | 0.183 | 0.306 | 0.561 |
| PPS | 79.198 | 97.813 | 141.035 |
| SDS | 5.687 | 5.772 | 5.939 |
| ECE | 32.415 | 55.695 | 447.576 |

The results in Table 2 demonstrate that the selected $\tau_k$ values effectively capture the characteristic scale of each component while remaining robust to extreme values. Notably, PPS and ECE present substantially larger scales compared to ARC and SDS, reflecting the inherently higher magnitude and variability of these components.

The use of the lower quartile ($\tau_{q25}$) as the reference scaling parameter ensures that moderate and high-performing values are not overly compressed during normalisation. This enables a more balanced contribution of each component to the final MADQI score while preserving sensitivity to meaningful variations in anomaly detection performance. Table 3 presents the normalised MADQI component scores computed on the test data.

**Table 3: Raw & Normalised Results**

| Metric | Raw Value | Normalized |
|---|---|---|
| ARC | 0.433 | 0.907 |
| PPS | 88.325 | 0.672 |
| SDS | 5.743 | 0.636 |
| ECE | 546.197 | 1.000 |

$MADQI = 0.25*ARC_{norm} + 0.25*PPS_{norm} + 0.25*SDS_{norm} + 0.25*ECE_{norm}$

$MADQI = 0.25 * 0.907 + 0.25 *672 + 0.25 *0.636 + 0.25 *1$

$MADQI = 0.8037$ → (80.37%)

The obtained MADQI score of approximately 80.37% indicates strong overall anomaly detection performance, suggesting that the model is effective in capturing meaningful anomalous behaviour in AIS data. To further illustrate the distribution of the normalised MADQI components, Figure 2 presents the component-wise performance profile of the model on the test dataset.

**Figure 2: MADQI Component Scores**

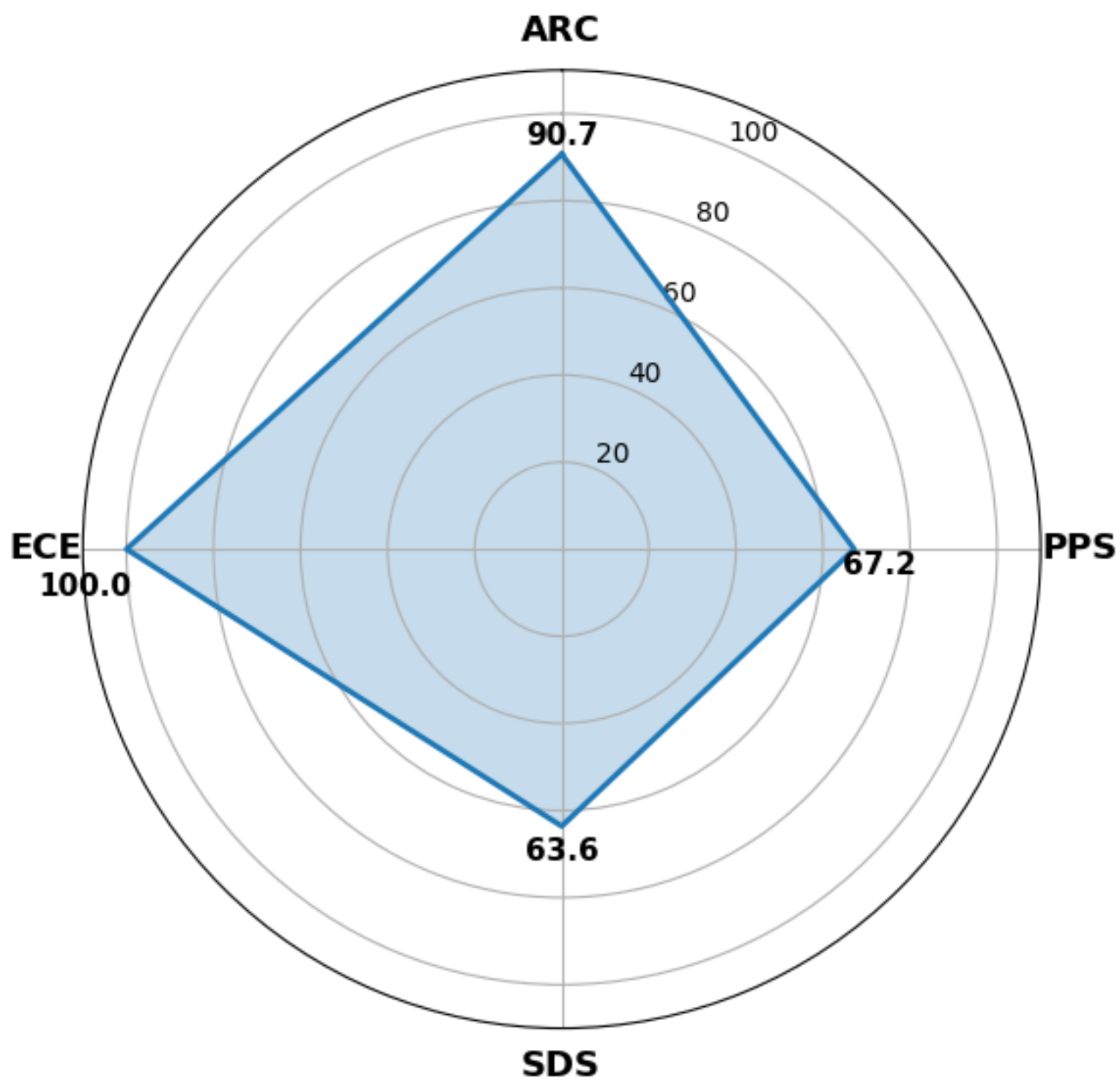

As shown in Figure 2, the model achieves its strongest performance in Extreme Case Evidence (ECE) and Anomaly Rate Consistency (ARC), while Physical Plausibility Score (PPS) and Score Distribution Separation (SDS) remain at moderate but meaningful levels.

The ARC value (0.907) indicates acceptable and stable consistency between observed and expected anomaly rates. This is complemented by the PPS value (0.672), which suggests that detected anomalies exhibit moderate to strong physical deviations from normal vessel behaviour. In parallel, the SDS value (0.636) reflects a clear separation between normal and anomalous observations in the anomaly score space. Most notably, the ECE value (1.000) demonstrates that the model effectively captures extreme behavioural deviations, particularly in spatial displacement.

## 5. DISCUSSION

The proposed MADQI framework provides a structured and interpretable approach for evaluating unsupervised anomaly detection models in maritime environments. Unlike conventional score-based assessments, MADQI enables a comprehensive understanding of model performance by integrating multiple complementary dimensions.

First, MADQI improves interpretability by decomposing anomaly detection performance into four meaningful components: consistency (ARC), physical plausibility (PPS), statistical separability (SDS), and extreme behavioural evidence (ECE). This component-wise structure allows practitioners to understand not only how well a model performs, but also the underlying characteristics of detected anomalies.

Second, the framework enables objective comparison across different models and configurations. By transforming heterogeneous metrics into a common bounded scale and aggregating them into a single composite index, MADQI facilitates consistent benchmarking without being affected by differences in metric magnitude or units.

Third, MADQI provides insight into model strengths and limitations. For example, a model may perform strongly in capturing extreme anomalies (high ECE) while exhibiting more moderate separation in score distributions (SDS). Such distinctions are particularly relevant in maritime monitoring, where different types of anomalies carry different operational importance.

In addition, the integration of multi-chunk evaluation and data-driven adaptive scaling improves the robustness of the framework. This reduces sensitivity to dataset-specific variations and avoids arbitrary parameter selection, leading to more reliable and stable evaluation outcomes.

Taken together, these characteristics make MADQI a practical and informative tool for assessing unsupervised anomaly detection performance in real-world maritime applications.

## 6. CONCLUSION AND CONTRIBUTIONS

This study presents a novel framework for maritime anomaly detection based on AIS data, integrating spatial analysis with unsupervised learning techniques. A central contribution of this work is the development of the Maritime Anomaly Detection Quality Index (MADQI), which addresses a key limitation in unsupervised anomaly detection by introducing a quantitative, scalable, and interpretable evaluation metric.

The proposed approach improves both the reliability and interpretability of anomaly detection systems, enabling more consistent assessment of model performance in real-world maritime monitoring contexts where labelled data are typically unavailable.

The main contributions of this study can be summarised as follows:

- **A novel composite evaluation metric (MADQI):** A model-agnostic and label-free index designed for evaluating unsupervised anomaly detection in maritime environments.
- **Multi-dimensional anomaly evaluation framework:** The integration of four complementary components such as Anomaly Rate Consistency (ARC), Physical Plausibility Score (PPS), Score Distribution Separation (SDS), and Extreme Case Evidence (ECE) to capture different aspects of anomaly detection performance.
- **Multi-chunk evaluation strategy:** A robustness-oriented approach that estimates metric distributions across multiple test subsets, reducing sensitivity to data-specific variations and improving evaluation stability.

- **Data-driven adaptive scaling mechanism:** Normalisation parameters are automatically derived from empirical distributions using lower-quantile statistics, eliminating arbitrary parameter selection.
- **Exponential normalisation formulation:** A novel transformation that avoids the centralisation bias of conventional soft normalisation and allows high-performing models to be more distinctly differentiated.
- **Fully unsupervised evaluation capability:** The framework operates without ground truth labels, addressing a fundamental limitation in maritime anomaly detection.
- **Scalable and domain-aware design:** The proposed method is computationally efficient and explicitly incorporates maritime-specific behavioural patterns, including vessel movement and spatial dynamics.

## 5. FUTURE WORK

Several directions can be explored to further extend the proposed MADQI framework. First, the applicability of MADQI can be evaluated across a wider range of anomaly detection models, including deep learning–based and hybrid approaches, to assess its generalisability beyond Isolation Forest.

Second, the framework can be enriched by incorporating multimodal data sources, such as satellite imagery, radar signals, and textual reports, enabling a more comprehensive understanding of maritime behaviour through multi-modal anomaly detection.

Third, although MADQI is designed for fully unsupervised settings, future studies may validate its effectiveness using expert-labelled datasets, providing additional empirical grounding and benchmarking opportunities.

Finally, the integration of MADQI into real-time maritime monitoring systems represents an important practical direction, allowing continuous assessment of anomaly detection performance in operational environments.

**Acknowledgement:** During the preparation of this work, the authors used ChatGPT (24 May 2023 version, OpenAI, San Francisco, CA, USA) to assist in language polishing, thereby improving the clarity and readability of the manuscript. After using this tool/service, the authors reviewed and edited the content as needed and took full responsibility for the content of the publication.

**Funding Statement:** This project has received funding from the European Union's Horizon Europe research and innovation programme under the Grant Agreement 101168489.

**Availability of Data and Materials:** In this study NOAA (2022) AIS dataset used to analyses. It is available on https://coast.noaa.gov/htdata/CMSP/AISDataHandler/2022/index.html. Additionally, dataset and Python Code notebook are available on https://zenodo.org/records/20425539.

**Ethics Approval:** Not applicable.

**Conflicts of Interest:** The authors declare no conflicts of interest.

## APPENDIX A: DERIVATION OF THE HAVERSINE FORMULA

### A.1. Geometric Intuition: The Chord and the Central Angle

To understand the haversine formulation, we first consider the relationship between the straight-line distance through the Earth (the chord) and the surface distance (the arc). Let two consecutive AIS positions on the Earth's surface be denoted by $A$ and $B$. Let the full central angle between two points be $2\theta$, so that the bisected angle in triangle AOC is $\theta$. This can be mapped onto a circle, as illustrated in Figure 1.

**Figure 1:** Geometric Representation of Haversine

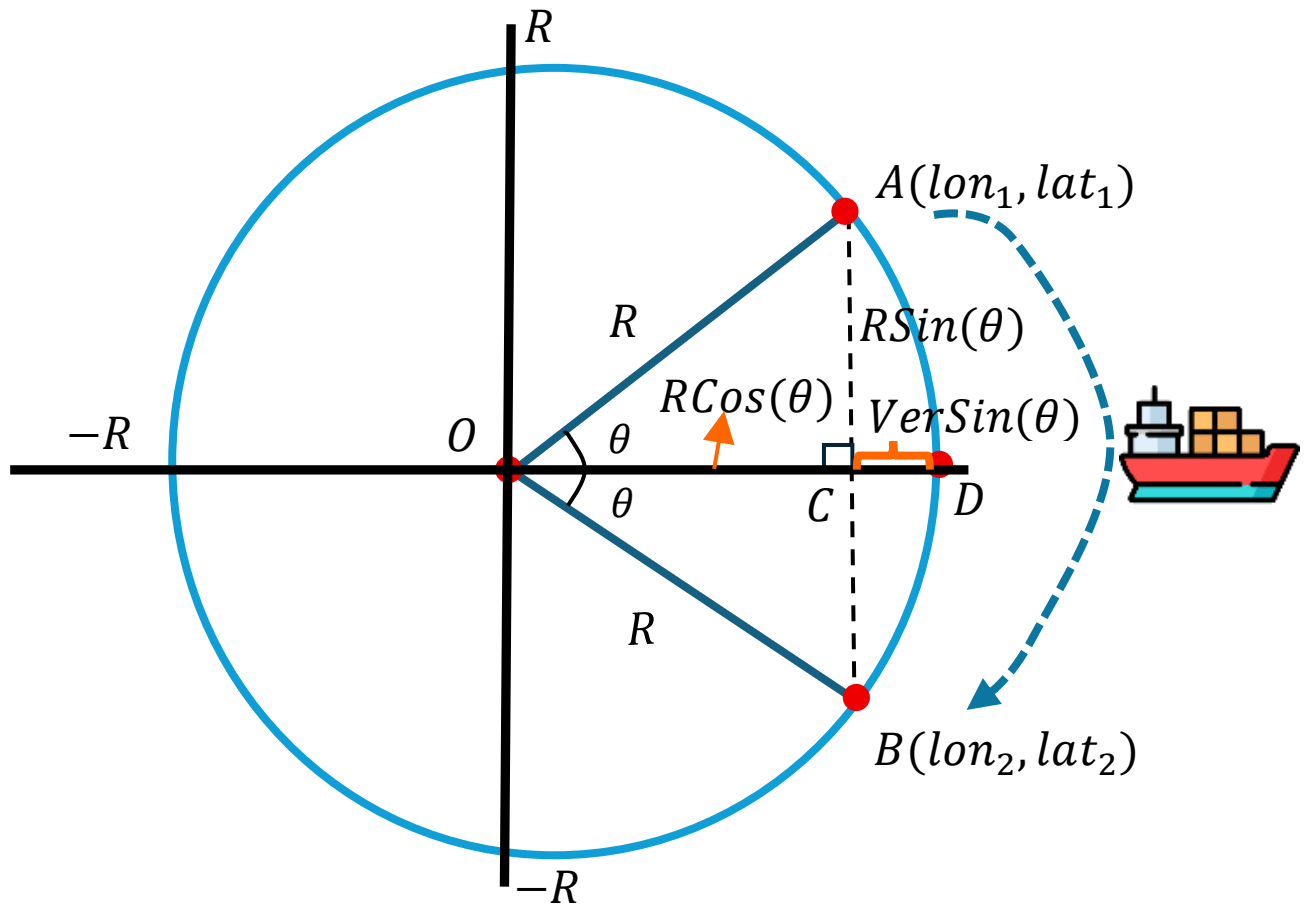

In Figure 1, the straight segment $|AB|$ represents the chord connecting the two positions, while the actual shortest path on the spherical surface is the great-circle arc. On a sphere, the surface distance $d$ is defined by the central angle as:

$$d = R(2\theta)$$

where $R$ is the Earth's mean radius. To build geometric intuition, let $C$ be the midpoint of chord $|AB|$. In the right-angled triangle $AOC$, the central angle is bisected, such that $AOC = \theta$. Therefore:

$$\sin(\theta) = \frac{|\,AC\,|}{R}$$

$$|\,AC\,| = \text{R}\sin(\theta)$$

Since $|AB| = 2|AC|$, the total chord length is given by:

$$|\,AB\,| = 2\text{R}\sin(\theta)$$

This geometric relationship explains why the half-angle term $(\theta)$ naturally appears in spherical distance calculations.

**A.2. The Haversine Function and the Unit Circle**

To simplify the trigonometric derivation, we map the triangle $AOC$ onto a unit circle $(R = 1)$. This allows us to define the **versine** (versed sine) and **haversine** (half-versed sine) functions.

As illustrated in Figure 1, let the distance from the centre to the chord be $|OC| = cos(\theta)$. The distance from the chord to the surface, $|CD|$, is known as the **versine**:

$$versin(\theta) = R - Rcos(\theta)$$

Using the trigonometric double-angle identity $cos(\alpha) = 1 - sin^2(\alpha/2)$, so $cos(\theta) = 1 - 2sin^2(\theta/2)$. We can rewrite this as:

$$versin(\theta) = 1 - \left(1 - 2 \cdot Sin^2(\theta/2)\right)$$

$$versin(\theta) = 2 \cdot Sin^2(\theta/2)$$

The **haversine** function is defined as half of the versine:

$$hav(\theta) = \frac{versin(\theta)}{2} = sin^2\left(\frac{\theta}{2}\right)$$

**A.3. Linking Angular Coordinates to Surface Distance**

In maritime navigation, the objective is to compute the great-circle distance $d$ using latitude $(lat)$ and longitude $(lon)$. According to the Spherical Law of Cosines, the central angle $\Delta\sigma$ between two points is related by:

$$cos(\Delta\sigma) = sin(lat_1)sin(lat_2) + cos(lat_1)cos(lat_2)cos(lon_2 - lon_1)$$

Using the identity $hav(x) = sin^2\left(\frac{x}{2}\right)$, the spherical law of cosines can be rewritten as:

$$sin^2\left(\frac{\Delta\sigma}{2}\right) = \sin^2\left(\frac{lat_2 - lat_1}{2}\right) + \cos(lat_1) \cdot \cos(lat_2) \cdot \sin^2\left(\frac{lon_2 - lon_1}{2}\right)$$

By substituting the haversine identity into this law, we obtain the relationship for the half-angle:

$$\sin^2\left(\frac{\Delta\sigma}{2}\right) = \sin^2\left(\frac{\Delta lat}{2}\right) + \cos(lat_1) \cdot \cos(lat_2) \cdot \sin^2\left(\frac{\Delta lon}{2}\right)$$

$$sin\left(\frac{\Delta\sigma}{2}\right) = \sqrt{\sin^2\left(\frac{\Delta lat}{2}\right) + \cos(lat_1) \cdot \cos(lat_2) \cdot \sin^2\left(\frac{\Delta lon}{2}\right)}$$

$$\frac{\Delta\sigma}{2} = arcsin\left(\sqrt{\sin^2\left(\frac{\Delta lat}{2}\right) + \cos(lat_1) \cdot \cos(lat_2) \cdot \sin^2\left(\frac{\Delta lon}{2}\right)}\right)$$

Since the arc length on a sphere is given by $d = R\Delta\sigma$, we obtain:

$$d = 2R \cdot arcsin\left(\sqrt{\sin^2\left(\frac{\Delta lat}{2}\right) + \cos(lat_1) \cdot \cos(lat_2) \cdot \sin^2\left(\frac{\Delta lon}{2}\right)}\right)$$

$$d = 2R \cdot arcsin\left(\sqrt{hav(\Delta lat) + \cos(lat_1) \cdot \cos(lat_2) hav(\Delta lon)}\right)$$